\documentclass[sigconf]{acmart}
\renewcommand\footnotetextcopyrightpermission[1]{} 
\AtBeginDocument{%
  \providecommand\BibTeX{{%
    \normalfont B\kern-0.5em{\scshape i\kern-0.25em b}\kern-0.8em\TeX}}}

\setcopyright{acmcopyright}
\copyrightyear{2020}
\acmYear{2020}
\acmDOI{10.1145/nnnnnnn.nnnnnnn}

\acmConference[ACM Multimedia]{}

\begin{document}
\newcommand\pjl[1]{\textcolor{black}{#1}}
\newcommand\gyy[1]{\textcolor{black}{#1}}

\title{Dense Scene Multiple Object Tracking with Box-Plane Matching}


\author{Jinlong Peng, Yueyang Gu, Yabiao Wang, Chengjie Wang, Jilin Li, Feiyue Huang}

\affiliation{%
\institution{Tencent Youtu Lab}
}
\email{{jeromepeng, yueyanggu, caseywang, jasoncjwang, jerolinli, garyhuang}@tencent.com}


\begin{abstract}
Multiple Object Tracking (MOT) is an important task in computer vision. MOT is still challenging due to the occlusion problem, especially in dense scenes. Following the tracking-by-detection framework, we propose the Box-Plane Matching (BPM) method to improve the MOT performacne in dense scenes. First, we design the Layer-wise Aggregation Discriminative Model (LADM) to filter the noisy detections. Then, to associate remaining detections correctly, we introduce the Global Attention Feature Model (GAFM) to extract appearance feature and use it to calculate the appearance similarity between history tracklets and current detections. Finally, we propose the Box-Plane Matching strategy to achieve data association according to the motion similarity and appearance similarity between tracklets and detections. With the effectiveness of the three modules, our team achieves the 1st place on the Track-1 leaderboard in the ACM MM Grand Challenge HiEve 2020.
\end{abstract}





\maketitle

\section{Introduction}
Multiple object tracking (MOT) is an important task in computer vision, which aims to generate the boxes and trajectories of the specific targets in the video~\cite{peng2020ctracker, zhu2018online}. Most exsiting MOT methods follow the tracking-by-detection framework, which obtain good tracking performance~\cite{tang2017multiple, peng2018tracklet, peng2020tpm}. However, MOT is still a challenging task due to the occlusion problem, especially in dense scenes. Severe occlusion could easily lead to target draft and trajectory interruption. And the numerous noisy detections and missed detections in dense scenes further increase the difficulty of tracking.

In order to solve the above problem, we propose a new method named Box-Plane Matching (BPM) for dense scene MOT based on Tracklet-Plane Matching (TPM) \cite{peng2020tpm}. First, we design the Layer-wise Aggregation Discriminative Model (LADM) to filter the noisy detections. Then, in order to associate remaining detections correctly, we introduce the Global Attention Feature Model (GAFM) to extract appearance feature and use it to calculate the appearance similarity between history tracklets and current detections. Finally, we propose the Box-Plane Matching strategy to achieve data association according to the motion similarity and appearance similarity between tracklets and detections.

The rest of the paper is introduced as follows. Three types of related work are reviewed in Section~\ref{sec:related_work}. Section~\ref{sec:algorithm} describes the details of our proposed BPM algorithm. The experimental results are presented in Section~\ref{sec:experiments}. Finally, Section~\ref{sec:conclusion} concludes the paper.

\section{Related Work}\label{sec:related_work}
Tracking-by-detection MOT methods can be converted to three steps: object detection, feature extraction and object association. Most of the researches focus on several of the three aspects to improve the MOT performance. In this paper, we propose Layer-wise Aggregation Discriminative Model, Global Attention Feature Model and Box-Plane Matching strategy, which are corresponding to the three aspects. Therefore, we introduce the related work according to these.

\textbf{Object detection}. Object detection plays an important role in MOT performance. Yu \textit{et.al} \cite{yu2016poi} proposed POI which conducted a high-performance detector by optimizing Faster R-CNN \cite{ren2015faster} and adding plenty of public pedestrian detection data. EDMT \cite{chen2017enhancing} designed an enhanced detection model which simultaneously modeled the detection-scene relation and detection-detection relation. Furthermore, Henschel \textit{et.al} \cite{Henschel2017} fused body detection and head detection to accomplish tracking, which needed extra training data and detection annotations.


\textbf{Feature extraction}. Robust appearance feature is critical for object association in MOT. LMP \cite{tang2017multiple} designed a novel person ReID model that combines the body pose layout. DMAN \cite{zhu2018online} used spatial and temporal attention mechanisms to extract the feature of tracklet and calculate the appearance similarity between history tracklet and current detection. Kim \textit{et.al} \cite{kim2018multi} proposed a bilinear LSTM based model to improve the long-term information feature learning via a recurrent network.

\textbf{Object association}. Object association is the core step of MOT, while both object detection and feature extraction are to assist it. Tang \textit{et.al} \cite{tang2015subgraph} proposed a Subgraph Multicut model to deal with object association. Furthermore, Yang \textit{et.al} \cite{yang2017hybrid} designed a min-cost multi-commodity network flow to fuse global optimization and local optimization in data association. Gao \textit{et.al} \cite{gao2017graphical} designed a Graphical Social Topology Model to improve the accuracy of detection association. TPM \cite{peng2020tpm} proposed global tracklet-plane matching which associated high-confidence short tracklets into long trajectories. However, when there are too many short tracklets caused by the long input videos and dense scenes, the speed of TPM will drop. Thus we proposed local box-plane matching to solve the dense scene MOT problem in this paper.

\begin{figure*}[t]
\centering
\includegraphics[width=0.9\textwidth]{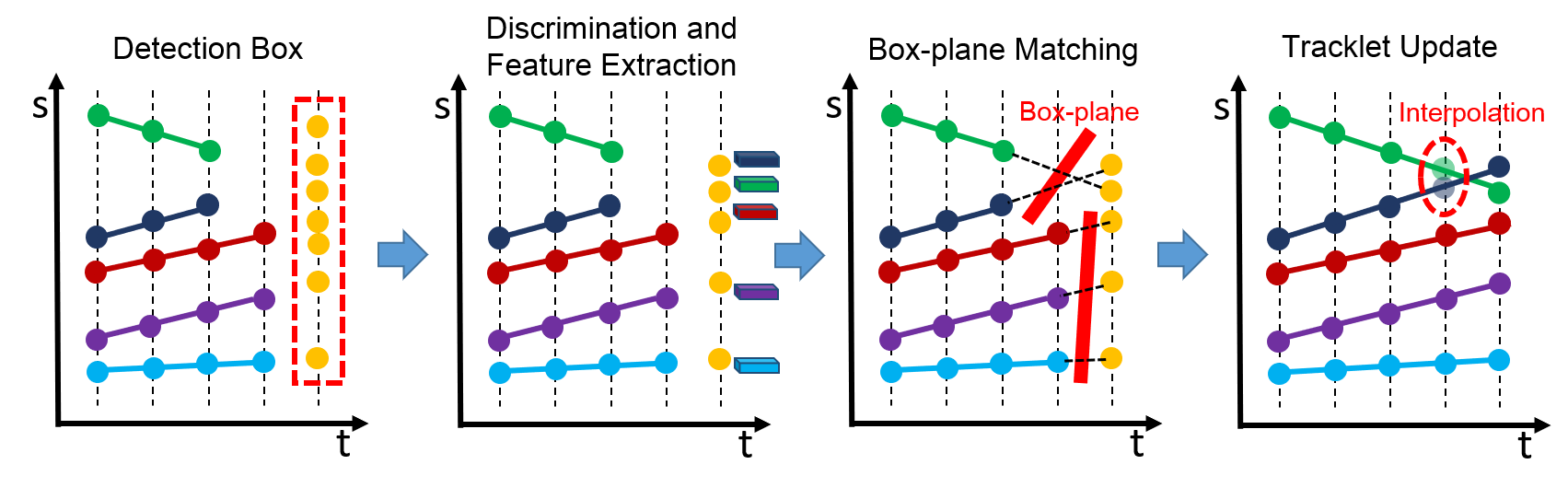}
\vspace{-10pt}
\caption{Illustration on the pipeline of BPM.}
\label{fig:BPM}
\end{figure*}

\section{Algorithm}\label{sec:algorithm}

In Track-1 of challenge HiEve 2020, contestants should use the public detections to accomplish tracking, which are provided by the host based on Faster R-CNN \cite{ren2015faster}. For the complicated and dense tracking scenes, we popose the Box-Plane Matching (BPM) method, whose framework is illustrated in Fig.~\ref{fig:BPM}. First, we design the Layer-wise Aggregation Discriminative Model (LADM) to filter the noisy detections in subsection \ref{subsection:LADM}. Then, we introduce the Global Attention Feature Model (GAFM) to extract appearance feature for the remaining detections in subsection \ref{subsection:GAFM}. Finally, we propose the Box-Plane Matching strategy to associate the remaining detections according to the similarity between history tracklets and current detections in subsection \ref{subsection:BPM}.

\subsection{Layer-wise Aggregation Discriminative Model\label{subsection:LADM}}

\pjl{We introduce the Layer-wise Aggregation Discriminative Model (LADM) to judge whether a detection box represents a person or not. LADM is actually a binary classification model. For every public detection box, we use it to crop the image from the located frame, which is the input of LADM.} 

\begin{figure}[t]
    \centering
    \includegraphics[width=\columnwidth]{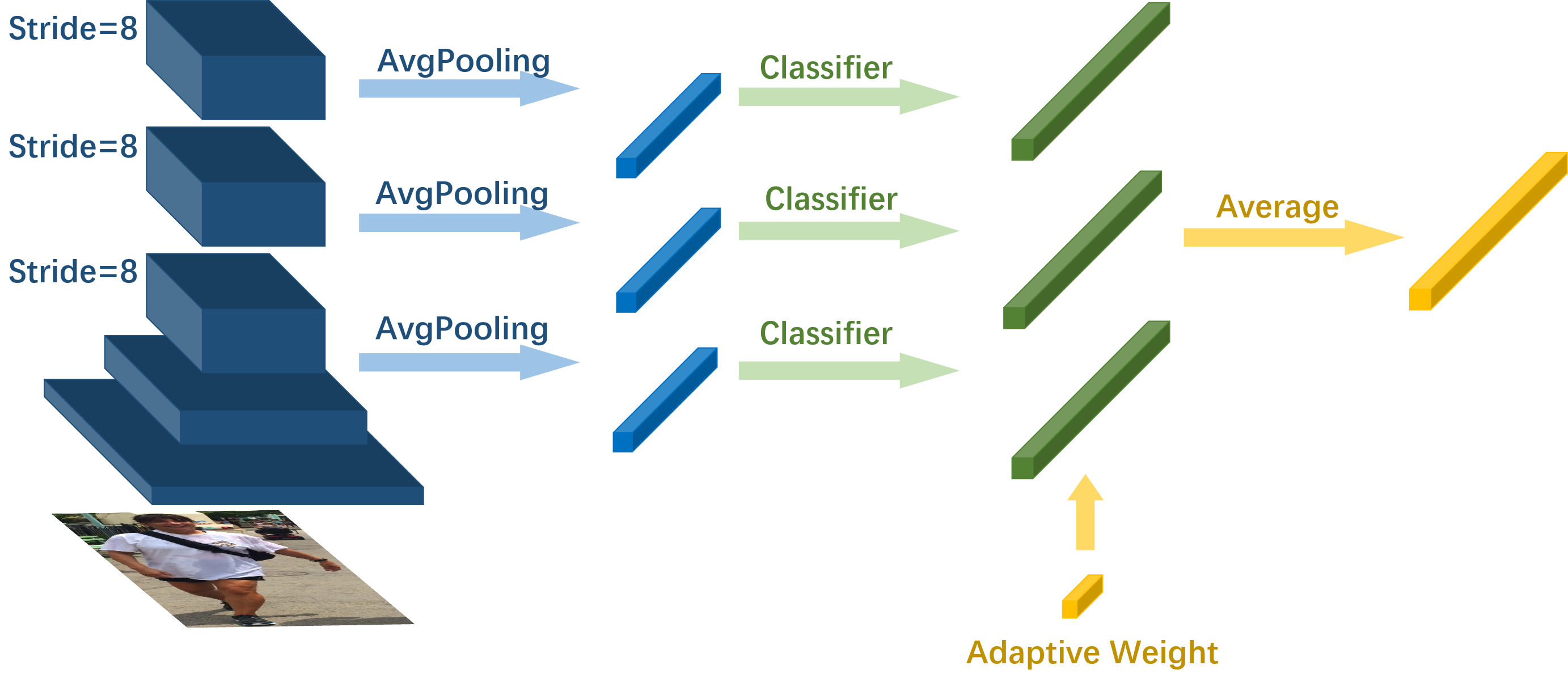}
    \caption{Structure of LADM.}
    \label{fig:LADM}
\end{figure}

The structure of LADM is displayed in Fig.~\ref{fig:LADM}. We utilize ResNet-152~\cite{ResNet} as backbone of LADM. Given that there is still a gap between training data and test data, LADM predicts whether a image is person or not based on layers-wise aggregation from three conv-feature maps. The final prediction is the weighted average of three softmax layer. These voting strategy from bagging can avoid over-fitting effectively.

The total stride of original ResNet152 is 32, which cause the resolution of final feature map too small. We adjust the stride of conv3 and conv4 from 2 to 1. This adjustment can bring two advantages:

\begin{itemize}
    \item The resolution of conv3 and conv4 are increased;
    \item The resolution of conv3, conv4 and conv5 are same.
\end{itemize}

The backbone outputs three conv-features of input image for classification. After pooling layer, they are input into classifier subnet. Each classifier subnet contains a fully connected layer and a softmax layer. The final prediction is the weighted average of three softmax layers. These weights are also parameters of LADM and optimized during training.

\subsection{Global Attention Feature Model\label{subsection:GAFM}}

\begin{figure}[t]
    \centering
    \includegraphics[width=\columnwidth]{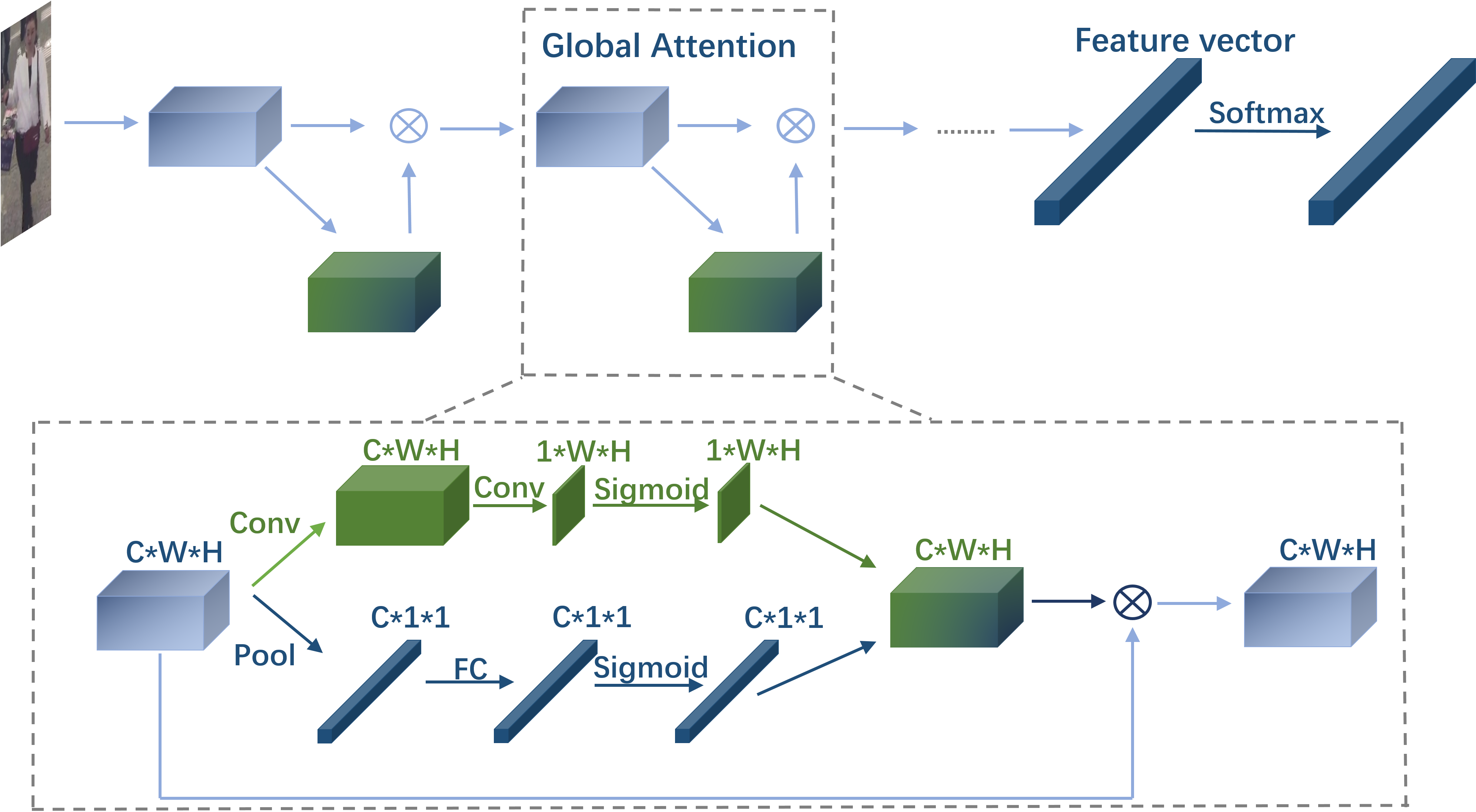}
    \caption{Structure of GAFM.}
    \label{fig:GAFM}
\end{figure}

\pjl{We design the Global Attention Feature Model (GAFM) to extract the appearance feature of the person that the detection box represents. GAFM can be seen as a person re-identification model. Each ID of person indicates one class. The structure of GAFM is illustrated in Fig.~\ref{fig:GAFM}.} We utilize ResNet152~\cite{ResNet} as backbone of GAFM. Since the view of different video sequences is different, the distribution of IDs are not same. We introduce global attention mechanism to enhance the discriminative representation of GAFM. 

Global attention mechanism is implemented by Global Attention Map (GAM). The shape of GAM is same as input feature map. Suppose that the shape of input feature map is C*W*H, the shape of GAM is also C*W*H. Global attention mechanism conducts element-wise multiplication between them and the result is attention feature map. This attention operation is integrated into residual branch of each bottleneck.

We decompose GAM as Spatial Attention Map (SAM) and Channel Attention Map (CAM). \gyy{SAM indicates spatial-aware information of targets while CAM encodes class-aware information about specific ID.} The shape of SAM is 1*W*H and of CAM is C*1*1. SAM is calculated as following:

\begin{itemize}
    \item Applying convolution with 3*3 kernel and keeping the number of channels unchanged. Then conducting point-wise convolution to reduce channels to 1. Sigmoid function here is used to normalized all the elements.
\end{itemize}

And CAM is calculated as following:

\begin{itemize}
    \item Applying global average pooling on input feature map. Then a fully connected layer is followed the pooling layer with same size. Sigmoid function here is used to normalized all the elements.
\end{itemize}

SAM and CAM are both repeated to C*W*H (same shape as input feature map). The final GAM is generated by simply combining SAM and CAM using element-wise multiplication.

\subsection{Box-Plane Matching\label{subsection:BPM}}

We propose the Box-Plane Matching (BPM) strategy based on TPM \cite{peng2020tpm}. TPM is used to associated high-confidence short tracklets into long trajectories in the global video, while BPM focuses on the local matching between history tracklets and current detection boxes, which is more suitable for long input videos and dense scenes. To construct the box-planes, we calculate the similarity between history tracklet $T$ and detection box $D$ as:
\begin{equation}
S(T,D)=A(T,D)+\lambda _{s}M(T,D),
\label{eq:similarity}
\end{equation}
where $S(T,D)$ represents the similarity between tracklet $T$ and detection box $D$. $A(T,D)$ is the appearance similarity and $M(T,D)$ is the motion similarity. $\lambda _s=1$ is the balance weight. We use the cosine similarity between the global attention feature as the appearance similarity, and the motion information of the tracklet and the box detection to calculate the motion similarity \cite{yu2016poi}.

Getting the similarity, we can construct the box-planes by the following optimization function:
{\begin{align}
&(X^*,Y^*,n_p^*)=\mathop{\arg\min}_{X,Y,n_p}\Phi _{1}(X,Y,n_p)+\Phi _{2}(X,Y,n_p)\nonumber,\\
&s.t.\quad x_{i}^{m},y_{j}^{m}\in\{0,1\};\sum_{m=1}^{n_{p}}x_{i}^{m}\leqslant 1;\sum_{m=1}^{n_{p}}y_{j}^{m}\leqslant 1
\label{eq:plane_optimization}
\end{align}}
where $X=\{x_{i}^{m}\}, i=1,...,n_t, m=1,...,n_p$ and $Y=\{y_{j}^{m}\}, j=1,...,n_d, m=1,...,n_p$ are the sets representing the box-plane construction status. $x_{i}^{m} = 1$ indicates the end of tracklet $T_i$ is connected to box-plane $P_m$ and $y_{j}^{m} = 1$ indicates detection box $D_j$ is connected to box-plane $P_m$. $n_t$ is the total number of tracklets, $n_d$ is the total number of detection boxes and $n_p$ is the total number of box-planes. The constraints guarantee that all the end of history tracklets and all the current detection boxes are connected to at most one box-plane. $\Phi _{1}(X,Y,n_p)$ and $\Phi _{2}(X,Y,n_p)$ are the optimization terms for evaluating box-plane construction qualities, which are defined in Eq.~\ref{eq:first_term} and Eq.~\ref{eq:second_term}, respectively:

\begin{equation}
\Phi _{1}(X,Y,n_p)=-\sum_{m=1}^{n_{p}}{\sum\limits_{i=1}^{n_{t}}\sum\limits_{j=1}^{n_{d}}2x_{i}^{m}y_{j}^{m}S(T_i,D_j)},
\label{eq:first_term}
\end{equation}
\begin{equation}
\Phi _{2}(X,Y,n_p)=\sum_{m=1}^{n_{p}}{\sum\limits_{i=1}^{n_{t}}\sum\limits_{j=1}^{n_{d}}(x_{i}^{m}x_{j}^{m}+y_{i}^{m}y_{j}^{m})S(T_i,D_j)},
\label{eq:second_term}
\end{equation}

$\Phi _{1}(X,Y,n_p)$ measures the total similarity between tracklets and detections connected to different sides of a tracklet-plane. While $\Phi _{2}(X,Y,n_p)$ measures the similarity separately among tracklets and among detections connected to the same side of a tracklet-plane. In this way, tracklets and detections representing the same objects are more likely to be connect to different sides of the same box-plane, while easily confusable tracklets or detections will not be connect to the same side of a box-plane, reducing mutual interference in the following in-plane matching step. Eqs.~\ref{eq:plane_optimization} can be solved by applying the Local Gradient Descent algorithm \cite{assari2016human}.

Then, we perform in-plane maching to update tracklets. The association of history tracklets and current detections is modeled as:
{\begin{align}
&Z^*=\mathop{\arg\max}_{Z}\sum_{m=1}^{n_{p}}\sum_{i=1}^{n_{t}}\sum_{j=1}^{n_{d}}x_{i}^{m}y_{j}^{m}z_{ij}S(T_i,D_j)\nonumber,\\
&s.t.\quad z_{ij}\in\{0,1\};\sum_{j=1}^{n_{d}}z_{ij}\leqslant 1;\sum_{i=1}^{n_{t}}z_{ij}\leqslant 1
\label{eq:inplane_optimization}
\end{align}}
where $Z=\{z_{ij}\}, i=1...n_t, j=1...n_d$ represents the association status of tracklets and detections. $z_{ij} = 1$ means the $D_j$ is connected to the end of the tracklet $T_i$. The constraints guarantee that every tracklet and detection box can only be connected once. We use the Kuhn-Munkres (KM) algorithm \cite{kuhn1955hungarian} applied to solve Eq.~\ref{eq:inplane_optimization}. Finally we apply the interpolation operation to generate complete and coherent trajectories.

\section{Experiments}\label{sec:experiments}
\subsection{Datasets and Evaluation Metrics}

HiEve2020 is a large-scale dataset for
human-centric video analysis \cite{lin2020human}. HiEve2020 contains 32 video sequences with 49820 frames. Since there are lots of dense scenes in the sequences, the amount of annotations is large, including 1302481 bounding boxes and 2687 track. Due to the diversity of data and the complexity of the scenes, the HiEve2020 test dataset can well reflect the accuracy and robustness of the MOT methods.

Moreover, we notice that the data distribution of training set and test set of HiEve2020 is quite different, especially for the density of persons. Thus, in addition to HiEve2020 training dataset, we add another extra dataset CrowdHuman \cite{shao2018crowdhuman} to establish training dataset for LADM. There are totally 15000 images and 339565 groundtruth bounding boxes in the CrowdHuman training datasets.

There are several metrics to evaluate the performance of MOT methods, including Multiple Object Tracking Accuracy (MOTA), Multiple Object Tracking Precision (MOTP), False Negatives (FN), False Positives (FP), Identity Switches (IDS), Mostly Tracked Trajectories (MT), Mostly Lost Trajectories (ML) \cite{bernardin2008evaluating, ristani2016performance}. Among these, MOTA is the primary metric to measure the overall MOT performance, thus the final performance ranking in HiEve2020 track-1 is based on MOTA. 

\subsection{Implementation Details}
\textbf{LADM: } We utilize groundtruth boxes in HiEve2020 and CrowdHuman training datasets to crop images as positive samples. We utilize random cropped image as negative samples, of which aspect ratio is from 0.5 to 4 and area is from 5000 to 50000. We notice that most of false positive detections in test dataset are background or partial body. Therefore, we crop negative samples based on the two principle: (1) The IoU between one negative sample and all groundtruth boxes in same image is smaller than 0.2; (2) The center of one negative sample is not in any groundtruth box. The training details of LADM is as following:

\begin{itemize}
    \item We introduce random cropping, random blurring, random erasing, horizontal flipping, gaussian normalization for data augmentation.
    \item We utilize pre-trained ResNet152 on ImageNet to initialize LADM. The learning rate of convolution layers is 0.1 times of classifier subnet.
    \item We use SGD with momentum as optimizer. We also utilize warm-up strategy and exponential learning rate schedule.
\end{itemize}

\noindent\textbf{GAFM: }In ground-truth of training dataset, each bounding-box is labelled with ID. These dataset can be used to train GAFM. Since there is little difference between two persons from adjacent frames, We select about 20 bounding-box per ID from interval frames to establish training dataset for GAFM. We exclude images with low resolution or severe occlusion. Finally, about 1100 IDs is used for training. The training details of LADM is as following:
\begin{itemize}
    \item We apply Instance Batch Normalization (IBN) into GAFM.
    \item We adjust the stride of final conv block from 2 to 1 for higher resolution feature map.
    \item We utilize random cropping, random erasing, horizontal flipping, gaussian normalization for data augmentation.
    \item We introduce Generalized Mean Pooling between backbone and fully connected layer.
    \item We use SGD with momentum as optimizer. We also utilize warm-up strategy and exponential learning rate schedule.
\end{itemize}

\begin{table}[t]
\renewcommand\arraystretch{1.2}
\centering
\caption{Ablation study on HiEve2020 validation dataset.}\label{tab:tab1} 
\vspace{-10pt}
\small{
\setlength{\tabcolsep}{1mm}{
\begin{tabular}{|c|ccccc|}
\hline
Method & MOTA$\uparrow$ & IDF1$\uparrow$ & MOTP$\uparrow$ & MT$\uparrow$ & ML$\downarrow$\\
\hline
Baseline & 47.7 & 45.8 & 76.1 & 26.8\% & 31.2\%\\
BP & 49.5 & 46.8 & 76.2 & 28.1\% & 30.0\%\\
BP+LADM & 51.2 & 47.3 & 76.7 & 28.9\% & 29.3\%\\
{\bf BP+LADM+GAFM (BPM)}& {\bf 51.8} & {\bf 47.5} & {\bf 76.8} & {\bf 29.7\%} & {\bf 28.8\%}\\

\hline
\end{tabular}}}
\end{table}

\subsection{Ablation Studies}

To prove the effectiveness of the different modules of our BPM methods, we conduct the following experiments:

(1) \textbf{Baseline.} Directly using the pretrain model of ResNet152 to extract feature, and applying the KM algorithm to associate the public detection boxes to generate trajectories.

(2) \textbf{BP.} Directly using the pretrain model of ResNet152 to extract feature, and applying the box-plane matching strategy to associate the public detection boxes.

(3) \textbf{BP+LADM.} First using LADM to filter noisy detections, then using the pretrain model of ResNet152 to extract feature, and applying the box-plane matching strategy to associate the public detection boxes.

(4) \textbf{BP+LADM+GAFM.} This the full version of our method BPM. First using LADM to filter noisy detections, then using GAFM to extract feature, and applying the box-plane matching strategy to associate the remaining detection boxes.

Table~\ref{tab:tab1} illustrates the results of the four methods on HiEve2020 validation dataset, selected from HiEve2020 training dataset. From Table~\ref{tab:tab1} we can find that:

(1) \textbf{BP} outperforms significantly than \textbf{Baseline}, which proves the effectiveness of the box-plane matching strategy. By applying the box-plane matching, the detection boxes can be associated more accurately, and the missed boxes can be interpolated correctly. Therefore, the MOTA increases from 47.7 to 49.5 and the IDF1 increases from 45.8 to 46.8.

(2) \textbf{BP+LADM} further outperforms \textbf{BP}, especially in MOTA, which increases from 49.5 to 51.2. By using LADM to filter noisy detections, the false positive detections decreased greatly and the difficulty of data association is reduced meanwhile. 

(3) \textbf{BP+LADM+GAFM} performs slightly better than \textbf{BP+LADM}. The MOTA is only improved by 0.6. Maybe in several sequences the images are blurry and the occlusion is serious, thus the appearance feature optimization has limited effect improvement.

\subsection{HiEve2020 Challenge Track-1 Results}

Table~\ref{tab:tab2} displays the Top-5 teams and their results in the ACM MM Grand Challenge HiEve2020 Track-1. Our team LinkBox achieves the 1st place on the leaderboard by using the full version of BPM, demonstrating the effectiveness of our method.

\begin{table}[t]
\renewcommand\arraystretch{1.2}
\centering
\caption{HiEve2020 Challenge Track-1 Results.}\label{tab:tab2} 
\vspace{-10pt}
\small{
\setlength{\tabcolsep}{1mm}{
\begin{tabular}{|c|c|c|}
\hline
Rank & Team & MOTA$\uparrow$\\
\hline
1 & {\bf LinkBox (Ours)} & 51.3828\\
2 & Selective JDE & 50.5863\\
3 & Try & 50.5518\\
4 & FCS-Track & 47.8048\\
5 & Commander & 47.4144\\

\hline
\end{tabular}}}
\end{table}

\section{Conclusion}\label{sec:conclusion}

In this paper we propose the Box-Plane Matching method to improve the MOT performacne in dense scenes. First, we design the Layer-wise Aggregation Discriminative Model to filter the noisy detections. Then we introduce the Global Attention Feature Model to extract appearance feature and use it to calculate the appearance similarity between history tracklets and current detections. Finally, we propose the Box-Plane Matching strategy to achieve data association according to the motion similarity and appearance similarity between tracklets and detections. We achieves the 1st place on the Track-1 leaderboard in the ACM MM Grand Challenge HiEve 2020, demonstrating the effectiveness of our approach.

\clearpage
\bibliographystyle{ACM-Reference-Format}
\bibliography{sample-base}


\begin{thebibliography}{20}


\ifx \showCODEN    \undefined \def \showCODEN     #1{\unskip}     \fi
\ifx \showDOI      \undefined \def \showDOI       #1{#1}\fi
\ifx \showISBNx    \undefined \def \showISBNx     #1{\unskip}     \fi
\ifx \showISBNxiii \undefined \def \showISBNxiii  #1{\unskip}     \fi
\ifx \showISSN     \undefined \def \showISSN      #1{\unskip}     \fi
\ifx \showLCCN     \undefined \def \showLCCN      #1{\unskip}     \fi
\ifx \shownote     \undefined \def \shownote      #1{#1}          \fi
\ifx \showarticletitle \undefined \def \showarticletitle #1{#1}   \fi
\ifx \showURL      \undefined \def \showURL       {\relax}        \fi
\providecommand\bibfield[2]{#2}
\providecommand\bibinfo[2]{#2}
\providecommand\natexlab[1]{#1}
\providecommand\showeprint[2][]{arXiv:#2}

\bibitem[\protect\citeauthoryear{Assari, Idrees, and Shah}{Assari
  et~al\mbox{.}}{2016}]%
        {assari2016human}
\bibfield{author}{\bibinfo{person}{Shayan~Modiri Assari},
  \bibinfo{person}{Haroon Idrees}, {and} \bibinfo{person}{Mubarak Shah}.}
  \bibinfo{year}{2016}\natexlab{}.
\newblock \showarticletitle{Human re-identification in crowd videos using
  personal, social and environmental constraints}. In
  \bibinfo{booktitle}{\emph{ECCV}}.
\newblock


\bibitem[\protect\citeauthoryear{Bernardin and Stiefelhagen}{Bernardin and
  Stiefelhagen}{2008}]%
        {bernardin2008evaluating}
\bibfield{author}{\bibinfo{person}{Keni Bernardin} {and}
  \bibinfo{person}{Rainer Stiefelhagen}.} \bibinfo{year}{2008}\natexlab{}.
\newblock \showarticletitle{Evaluating multiple object tracking performance:
  the CLEAR MOT metrics}.
\newblock \bibinfo{journal}{\emph{EURASIP Journal on Image and Video
  Processing}} (\bibinfo{year}{2008}).
\newblock


\bibitem[\protect\citeauthoryear{Chen, Sheng, Zhang, and Xiong}{Chen
  et~al\mbox{.}}{2017}]%
        {chen2017enhancing}
\bibfield{author}{\bibinfo{person}{Jiahui Chen}, \bibinfo{person}{Hao Sheng},
  \bibinfo{person}{Yang Zhang}, {and} \bibinfo{person}{Zhang Xiong}.}
  \bibinfo{year}{2017}\natexlab{}.
\newblock \showarticletitle{Enhancing detection model for multiple hypothesis
  tracking}. In \bibinfo{booktitle}{\emph{CVPRW}}.
\newblock


\bibitem[\protect\citeauthoryear{Gao, Chen, Ye, Xing, Kuijper, and Ji}{Gao
  et~al\mbox{.}}{2017}]%
        {gao2017graphical}
\bibfield{author}{\bibinfo{person}{Shan Gao}, \bibinfo{person}{Xiaogang Chen},
  \bibinfo{person}{Qixiang Ye}, \bibinfo{person}{Junliang Xing},
  \bibinfo{person}{Arjan Kuijper}, {and} \bibinfo{person}{Xiangyang Ji}.}
  \bibinfo{year}{2017}\natexlab{}.
\newblock \showarticletitle{Beyond Group: Multiple Person Tracking via Minimal
  Topology-Energy-Variation}.
\newblock \bibinfo{journal}{\emph{IEEE TIP}} (\bibinfo{year}{2017}).
\newblock


\bibitem[\protect\citeauthoryear{He, Zhang, Ren, and Sun}{He
  et~al\mbox{.}}{2016}]%
        {ResNet}
\bibfield{author}{\bibinfo{person}{Kaiming He}, \bibinfo{person}{Xiangyu
  Zhang}, \bibinfo{person}{Shaoqing Ren}, {and} \bibinfo{person}{Jian Sun}.}
  \bibinfo{year}{2016}\natexlab{}.
\newblock \showarticletitle{Deep Residual Learning for Image Recognition}. In
  \bibinfo{booktitle}{\emph{CVPR}}.
\newblock


\bibitem[\protect\citeauthoryear{Henschel, Leal-Taix{\'e}, Cremers, and
  Rosenhahn}{Henschel et~al\mbox{.}}{2018}]%
        {Henschel2017}
\bibfield{author}{\bibinfo{person}{Roberto Henschel}, \bibinfo{person}{Laura
  Leal-Taix{\'e}}, \bibinfo{person}{Daniel Cremers}, {and}
  \bibinfo{person}{Bodo Rosenhahn}.} \bibinfo{year}{2018}\natexlab{}.
\newblock \showarticletitle{Fusion of head and full-body detectors for
  multi-object tracking}. In \bibinfo{booktitle}{\emph{CVPRW}}.
\newblock


\bibitem[\protect\citeauthoryear{Kim, Li, and Rehg}{Kim et~al\mbox{.}}{2018}]%
        {kim2018multi}
\bibfield{author}{\bibinfo{person}{Chanho Kim}, \bibinfo{person}{Fuxin Li},
  {and} \bibinfo{person}{James~M Rehg}.} \bibinfo{year}{2018}\natexlab{}.
\newblock \showarticletitle{Multi-object Tracking with Neural Gating Using
  Bilinear LSTM}. In \bibinfo{booktitle}{\emph{ECCV}}.
\newblock


\bibitem[\protect\citeauthoryear{Kuhn}{Kuhn}{1955}]%
        {kuhn1955hungarian}
\bibfield{author}{\bibinfo{person}{Harold~W Kuhn}.}
  \bibinfo{year}{1955}\natexlab{}.
\newblock \showarticletitle{The Hungarian method for the assignment problem}.
\newblock \bibinfo{journal}{\emph{NRL}} (\bibinfo{year}{1955}).
\newblock


\bibitem[\protect\citeauthoryear{Lin, Liu, Liu, Li, Qi, Qian, Wang, Sebe, Xu,
  Xiong, et~al\mbox{.}}{Lin et~al\mbox{.}}{2020}]%
        {lin2020human}
\bibfield{author}{\bibinfo{person}{Weiyao Lin}, \bibinfo{person}{Huabin Liu},
  \bibinfo{person}{Shizhan Liu}, \bibinfo{person}{Yuxi Li},
  \bibinfo{person}{Guo-Jun Qi}, \bibinfo{person}{Rui Qian},
  \bibinfo{person}{Tao Wang}, \bibinfo{person}{Nicu Sebe},
  \bibinfo{person}{Ning Xu}, \bibinfo{person}{Hongkai Xiong}, {et~al\mbox{.}}}
  \bibinfo{year}{2020}\natexlab{}.
\newblock \showarticletitle{Human in Events: A Large-Scale Benchmark for
  Human-centric Video Analysis in Complex Events}.
\newblock \bibinfo{journal}{\emph{arXiv preprint arXiv:2005.04490}}
  (\bibinfo{year}{2020}).
\newblock


\bibitem[\protect\citeauthoryear{Peng, Qiu, See, Guo, Huang, Duan, and
  Lin}{Peng et~al\mbox{.}}{2018}]%
        {peng2018tracklet}
\bibfield{author}{\bibinfo{person}{Jinlong Peng}, \bibinfo{person}{Fan Qiu},
  \bibinfo{person}{John See}, \bibinfo{person}{Qi Guo},
  \bibinfo{person}{Shaoshuai Huang}, \bibinfo{person}{Ling-Yu Duan}, {and}
  \bibinfo{person}{Weiyao Lin}.} \bibinfo{year}{2018}\natexlab{}.
\newblock \showarticletitle{Tracklet Siamese Network with Constrained
  Clustering for Multiple Object Tracking}. In
  \bibinfo{booktitle}{\emph{VCIP}}.
\newblock


\bibitem[\protect\citeauthoryear{Peng, Wang, Wan, Wu, Wang, Tai, Wang, Li,
  Huang, and Fu}{Peng et~al\mbox{.}}{2020b}]%
        {peng2020ctracker}
\bibfield{author}{\bibinfo{person}{Jinlong Peng}, \bibinfo{person}{Changan
  Wang}, \bibinfo{person}{Fangbin Wan}, \bibinfo{person}{Yang Wu},
  \bibinfo{person}{Yabiao Wang}, \bibinfo{person}{Ying Tai},
  \bibinfo{person}{Chengjie Wang}, \bibinfo{person}{Jilin Li},
  \bibinfo{person}{Feiyue Huang}, {and} \bibinfo{person}{Yanwei Fu}.}
  \bibinfo{year}{2020}\natexlab{b}.
\newblock \showarticletitle{Chained-Tracker: Chaining Paired Attentive
  Regression Results for End-to-End Joint Multiple-Object Detection and
  Tracking}. In \bibinfo{booktitle}{\emph{ECCV}}.
\newblock


\bibitem[\protect\citeauthoryear{Peng, Wang, Lin, Wang, See, Wen, and
  Ding}{Peng et~al\mbox{.}}{2020a}]%
        {peng2020tpm}
\bibfield{author}{\bibinfo{person}{Jinlong Peng}, \bibinfo{person}{Tao Wang},
  \bibinfo{person}{Weiyao Lin}, \bibinfo{person}{Jian Wang},
  \bibinfo{person}{John See}, \bibinfo{person}{Shilei Wen}, {and}
  \bibinfo{person}{Erui Ding}.} \bibinfo{year}{2020}\natexlab{a}.
\newblock \showarticletitle{TPM: Multiple Object Tracking with Tracklet-Plane
  Matching}.
\newblock \bibinfo{journal}{\emph{Pattern Recognition}} (\bibinfo{year}{2020}).
\newblock


\bibitem[\protect\citeauthoryear{Ren, He, Girshick, and Sun}{Ren
  et~al\mbox{.}}{2015}]%
        {ren2015faster}
\bibfield{author}{\bibinfo{person}{Shaoqing Ren}, \bibinfo{person}{Kaiming He},
  \bibinfo{person}{Ross Girshick}, {and} \bibinfo{person}{Jian Sun}.}
  \bibinfo{year}{2015}\natexlab{}.
\newblock \showarticletitle{Faster r-cnn: Towards real-time object detection
  with region proposal networks}. In \bibinfo{booktitle}{\emph{NIPS}}.
\newblock


\bibitem[\protect\citeauthoryear{Ristani, Solera, Zou, Cucchiara, and
  Tomasi}{Ristani et~al\mbox{.}}{2016}]%
        {ristani2016performance}
\bibfield{author}{\bibinfo{person}{Ergys Ristani}, \bibinfo{person}{Francesco
  Solera}, \bibinfo{person}{Roger Zou}, \bibinfo{person}{Rita Cucchiara}, {and}
  \bibinfo{person}{Carlo Tomasi}.} \bibinfo{year}{2016}\natexlab{}.
\newblock \showarticletitle{Performance measures and a data set for
  multi-target, multi-camera tracking}. In \bibinfo{booktitle}{\emph{ECCV}}.
\newblock


\bibitem[\protect\citeauthoryear{Shao, Zhao, Li, Xiao, Yu, Zhang, and Sun}{Shao
  et~al\mbox{.}}{2018}]%
        {shao2018crowdhuman}
\bibfield{author}{\bibinfo{person}{Shuai Shao}, \bibinfo{person}{Zijian Zhao},
  \bibinfo{person}{Boxun Li}, \bibinfo{person}{Tete Xiao},
  \bibinfo{person}{Gang Yu}, \bibinfo{person}{Xiangyu Zhang}, {and}
  \bibinfo{person}{Jian Sun}.} \bibinfo{year}{2018}\natexlab{}.
\newblock \showarticletitle{Crowdhuman: A benchmark for detecting human in a
  crowd}.
\newblock \bibinfo{journal}{\emph{arXiv preprint arXiv:1805.00123}}
  (\bibinfo{year}{2018}).
\newblock


\bibitem[\protect\citeauthoryear{Tang, Andres, Andriluka, and Schiele}{Tang
  et~al\mbox{.}}{2015}]%
        {tang2015subgraph}
\bibfield{author}{\bibinfo{person}{Siyu Tang}, \bibinfo{person}{Bjoern Andres},
  \bibinfo{person}{Mykhaylo Andriluka}, {and} \bibinfo{person}{Bernt Schiele}.}
  \bibinfo{year}{2015}\natexlab{}.
\newblock \showarticletitle{Subgraph decomposition for multi-target tracking}.
  In \bibinfo{booktitle}{\emph{CVPR}}.
\newblock


\bibitem[\protect\citeauthoryear{Tang, Andriluka, Andres, and Schiele}{Tang
  et~al\mbox{.}}{2017}]%
        {tang2017multiple}
\bibfield{author}{\bibinfo{person}{Siyu Tang}, \bibinfo{person}{Mykhaylo
  Andriluka}, \bibinfo{person}{Bjoern Andres}, {and} \bibinfo{person}{Bernt
  Schiele}.} \bibinfo{year}{2017}\natexlab{}.
\newblock \showarticletitle{Multiple people tracking by lifted multicut and
  person re-identification}. In \bibinfo{booktitle}{\emph{CVPR}}.
\newblock


\bibitem[\protect\citeauthoryear{Yang, Wu, and Jia}{Yang et~al\mbox{.}}{2017}]%
        {yang2017hybrid}
\bibfield{author}{\bibinfo{person}{Min Yang}, \bibinfo{person}{Yuwei Wu}, {and}
  \bibinfo{person}{Yunde Jia}.} \bibinfo{year}{2017}\natexlab{}.
\newblock \showarticletitle{A Hybrid Data Association Framework for Robust
  Online Multi-Object Tracking}.
\newblock \bibinfo{journal}{\emph{IEEE TIP}} (\bibinfo{year}{2017}).
\newblock


\bibitem[\protect\citeauthoryear{Yu, Li, Li, Liu, Shi, and Yan}{Yu
  et~al\mbox{.}}{2016}]%
        {yu2016poi}
\bibfield{author}{\bibinfo{person}{Fengwei Yu}, \bibinfo{person}{Wenbo Li},
  \bibinfo{person}{Quanquan Li}, \bibinfo{person}{Yu Liu},
  \bibinfo{person}{Xiaohua Shi}, {and} \bibinfo{person}{Junjie Yan}.}
  \bibinfo{year}{2016}\natexlab{}.
\newblock \showarticletitle{POI: multiple object tracking with high performance
  detection and appearance feature}. In \bibinfo{booktitle}{\emph{ECCV}}.
\newblock


\bibitem[\protect\citeauthoryear{Zhu, Yang, Liu, Kim, Zhang, and Yang}{Zhu
  et~al\mbox{.}}{2018}]%
        {zhu2018online}
\bibfield{author}{\bibinfo{person}{Ji Zhu}, \bibinfo{person}{Hua Yang},
  \bibinfo{person}{Nian Liu}, \bibinfo{person}{Minyoung Kim},
  \bibinfo{person}{Wenjun Zhang}, {and} \bibinfo{person}{Ming-Hsuan Yang}.}
  \bibinfo{year}{2018}\natexlab{}.
\newblock \showarticletitle{Online multi-object tracking with dual matching
  attention networks}. In \bibinfo{booktitle}{\emph{ECCV}}.
\newblock


\end{thebibliography}

\end{document}